\documentclass[11pt]{article}

\usepackage{courier}
\usepackage{hyperref}
\usepackage{graphicx,subfigure}
\usepackage{epstopdf}
\usepackage{tikz}
\usepackage{amsmath}
\usepackage{multirow}
\usepackage{amsfonts}
\usepackage{float}
\usepackage{amssymb}
\usepackage{graphicx}
\usepackage{flushend}

\usepackage[linesnumbered,ruled,vlined]{algorithm2e}
\makeatletter

\newcommand{\Rmnum}[1]{\expandafter\@slowromancap\romannumeral #1@}

\makeatother

\AtBeginDocument{%
  \let\oldref\ref%
  \def\ref{\oldref*}}
  
\begin{document}
\title{\LARGE Deep Reinforcement Learning with Mixed Convolutional Network}
\author{Yanyu Zhang and Hairuo Sun}
\maketitle
\let\thefootnote\relax\footnotetext{\noindent\underbar{\hspace{0.8in}}\\
Yanyu Zhang is with the Departments of Electrical and Computer Engineering at Boston University, Boston, MA 02115. May be reached at {\tt zhangya@bu.edu}.}

\begin{abstract}
Recent research has shown that map raw pixels from a single front-facing camera directly to steering commands are surprisingly powerful (\cite{Bojarski}). This paper presents a convolutional neural network (CNN) \cite{Pomerleau} to playing the CarRacing-v0 using imitation learning in OpenAI Gym \cite{Brockman}. The dataset is generated by playing the game manually in Gym and used a data augmentation method to expand the dataset to 4 times larger than before. Also, we read the true speed, four ABS sensors, steering wheel position, and gyroscope for each image and designed a mixed model \cite{Rodriguez} by combining the sensor input and image input. After training, this model can automatically detect the boundaries of road features and drive the robot like a human. By comparing with AlexNet \cite{Krizhevsky} and VGG16 \cite{Simonyan} using the average reward in CarRacing-v0, our model wins the maximum overall system performance.
\end{abstract}

\section{Introduction}
OpenAI Gym is a popular open-source repository of reinforcement learning (RL) (\cite{Sutton} \cite{Kaelbling}) environments and development tools. Its curated set of problems and ease of use have made it a standard benchmarking tool for RL algorithms. We saw OpenAI Gym as an ideal tool for venturing deeper into RL. Our objective was to conquer an RL problem far closer to real-world use cases and the CarRacing-v0 environment provided exactly this.

In this paper, we generated a dataset with over 10,000 sample pairs and each sample pair contains one RGB image and three corresponding output control actions. Then we used a data augmentation method (\cite{Tanner}) to expand the dataset to 4 times larger than before and compressed the three channels RGB image into one channel grayscale image. In order to improve the robustness of our model, we combined some sensing elements \cite{Verma} by obtaining the immediate true speed, four ABS sensors, steering wheel position, and gyroscope from the CarRacing-v0 system. After that, we trained a mixed convolutional neural network and compared the performance with AlexNet and VGG16 in the same environment in the Gym.

\section{Data Preparation}
The dataset was generated by playing the game manually in the Gym. To increase the quality of the dataset, we made three changes to make it easier to control the robot. Firstly, we increased break strength, which allowed us to generally drive faster and reduce the velocity just before turns. Secondly, implement logic that automatically cuts gas when the turning, which largely reduced drifting. Thirdly, we adjust the line speed and spinning speed to make it better to control.

\begin{figure}[H]
	\centering
	\subfigure[ ]{
		\label{Fig.sub.1}
		\includegraphics[scale=0.5]{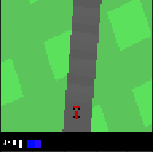}}
	\subfigure[ ]{
		\label{Fig.sub.2}
		\includegraphics[scale=0.5]{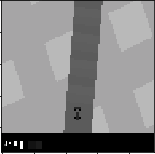}}
	\subfigure[ ]{
		\label{Fig.sub.3}
		\includegraphics[scale=0.5]{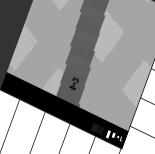}}
	\subfigure[ ]{
		\label{Fig.sub.4}
		\includegraphics[scale=0.5]{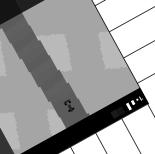}}
	\subfigure[ ]{
		\label{Fig.sub.5}
		\includegraphics[scale=0.5]{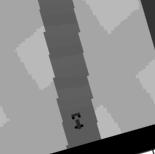}}
	\caption{Data augmentation processing.  {\bf (a)} The original RGB image.  {\bf (b)} The grayscale image. {\bf (c) (d) (e)} Data augmentation by randomly cropping, padding, and horizontal flipping. }
	\label{fig:yyz:aug}
\end{figure}

Before using the expert data to train the network, we did some data preprocessing. Firstly, we used a data augmentation method by randomly cropping, padding, and horizontal flipping for each image and expand the dataset to 4 times larger than before in Fig.\ \ref{fig:yyz:aug}. Then we used an image compression method (\cite{Rabbani}) by converting an image with RGB channels into an image with a single grayscale channel. But not all samples from the expert dataset were used for training. We started by discarding the first 50 samples of the episode. Those samples are peculiar because they show the track from a different zoom level. Therefore, there is a profit, even that small, on removing this heterogeneity from the training dataset. With this rule, the agent will never encounter states with unusual zooming levels.

\begin{figure}[H]
	\centering
	\subfigure[ ]{
		\label{Fig.sub.21}
		\includegraphics[scale=0.6]{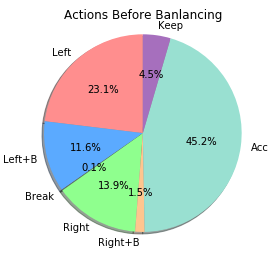}}
	\subfigure[ ]{
		\label{Fig.sub.22}
		\includegraphics[scale=0.6]{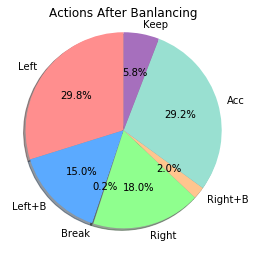}}
	\caption{{\bf (a)} Actions distribution before balancing.  {\bf (b)} Actions distribution After balancing.}
	\label{fig:yyz:dist}
\end{figure}

After that, we discarded samples in order to have a more balanced distribution of actions. Fig.\ \ref{fig:yyz:dist} (a) shows the original distribution of actions in the dataset. The label of each action is shown in Table.\ \ref{tab:1}. As we can see, there is a major predominance of the action accelerate. This may cause problems while training the network, because it may converge to the solution of always outputting accelerate. So we randomly removed 50\% of the samples labeled as accelerates. The resulting data distribution is shown in Fig.\ \ref{fig:yyz:dist} (b).

\begin{table}[h]
	\caption{Labels and Actions}    
	\centering
	\begin{tabular}{cc}
		\hline
		Label & Action \\  
		\hline
		Left & Turn left\\ 
		Left+B & Turn left with breaking\\ 
		Right & Turn right\\ 
		Right+B & Turn right with breaking\\ 
		Keep & Keep forward\\
		Acc & Accelerate\\
		Break & Brake\\
		\hline
	\end{tabular}
	\label{tab:1}
\end{table}

\section{Mixed Convolutional Neural Network}
\begin{figure}[h]
	\begin{center}
		\includegraphics[scale=0.5]{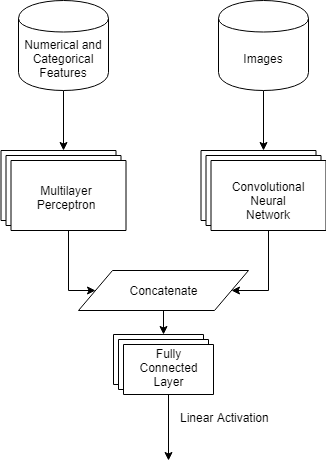}
	\end{center}
	\caption{Neural network structure, numerical features input are the feedback from the Gym sensors, images input are coming from the front-facing camera on robot.}
	\label{fig:yyz:model}
\end{figure}

The most important part of the agent is the neural network, which was totally written in PyTorch \cite{Paszke}. We designed a mixed neural network by combining robotics sensing features and images in Fig.\ \ref{fig:yyz:model}. The sensing feature can be seen as seven numerics for each corresponding image and we used a full connection layer in the perception section. On the other way,  the architecture of our CNN model is shown in Table.\ \ref{tab:2}. We used Adam with a learning rate of $10^{-5}$ as the optimizer for minimizing a cross-entropy loss \cite{Shore}. The network was trained for 100 epochs with batches of 64 samples. 

The concatenation appeared after the two convolutional layers in the CNN model, and the numeric input downside to one dimension and added into the flatten matrix at the ID 5 in Table.\ \ref{tab:2}. Then we used two linear layers to decrease the features as same as output we need. In this way, we reasonably combined sensor variables and the image.

\begin{table}[h]
	\caption{Neural Network}    
	\centering
	\begin{tabular}{ccccc}
		\hline
		ID & Type & Size & Stride & Activ. \\  
		\hline
		1 & Conv2D & 5x5x16 filter & 4x4 & ReLU\\ 
		2 & Dropout & 20\% Drop & -- & --\\ 
		3 & Conv2D & 3x3x32 filter & 2x2 & ReLU\\ 
		4 & Dropout & 20\% Drop & -- & --\\ 
		5 & Flatten & (Sensor Input) & -- & --\\ 
		6 & Dense & 128 Units & -- & Linear\\ 
		7 & Dense & 7 Units & -- & Softmax\\ 
		\hline
	\end{tabular}
	\label{tab:2}
\end{table}

\section{Results and Discussion}
\begin{figure}[H]
	\centering
	\subfigure[ ]{
		\label{Fig.sub.31}
		\includegraphics[scale=0.3]{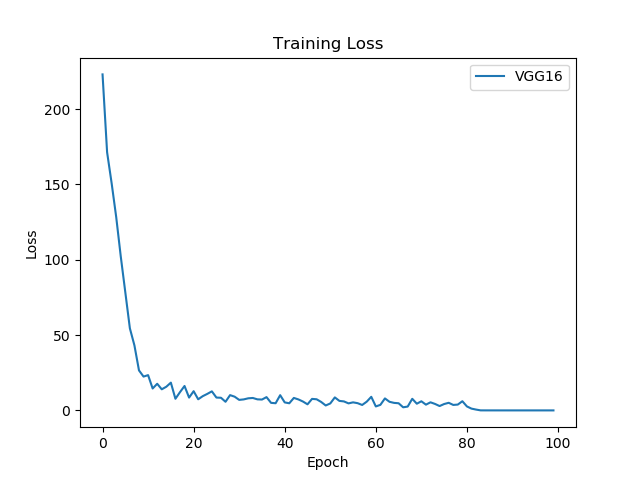}}
	\subfigure[ ]{
		\label{Fig.sub.32}
		\includegraphics[scale=0.3]{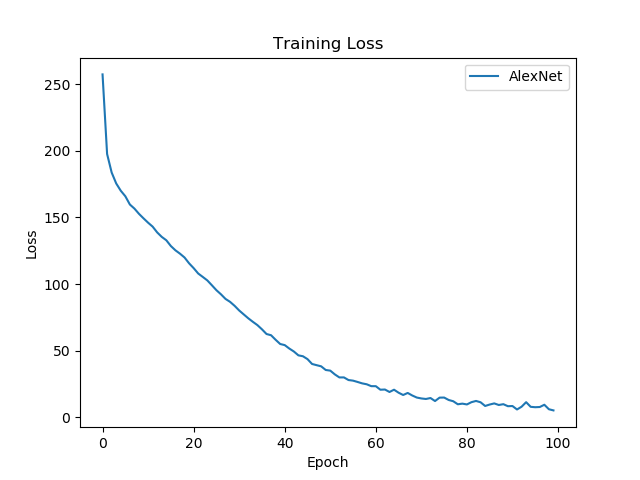}}
	\subfigure[ ]{
		\label{Fig.sub.33}
		\includegraphics[scale=0.3]{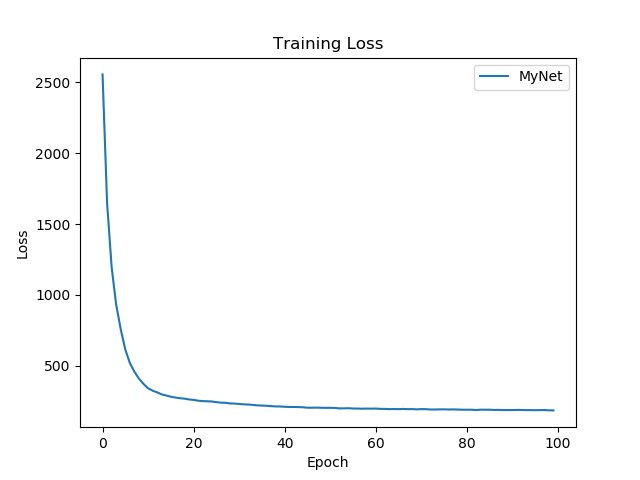}}
	\caption{{\bf (a)} VGG16.  {\bf (b)} AlexNet. {\bf (c)} MyNet.}
	\label{fig:yyz:result}
\end{figure}

After training on one NVIDIA 2080Ti GPU, the loss throughout the epochs is shown in Fig.\ \ref{fig:yyz:result}. As we can see, the loss for VGG16 and MyNet becomes stabilizes early than AlexNet. However, the loss for MyNet is higher than the others. Then we used the agent to test the final performance to control a robot in Gym. We tested 10 episodes for each network and took the average award for them, which recorded in Table.\ \ref{tab:3}.

\begin{table}[h]
	\caption{Average Score Running in Gym Agent}    
	\centering
	\begin{tabular}{cccc}
		\hline
		NN & VGG16 & AlexNet & MyNet \\  
		\hline
		RGB & 438.8 & 471.5 & 594.9\\ 
		Grayscale & 432.4 & 464.9 & 558.3\\ 
		\hline
	\end{tabular}
	\label{tab:3}
\end{table}

From the Table.\ \ref{tab:3}, we can see our mixed convolutional neural network earns the highest score in both the three-channel model and grayscale model. Also, we find three-channel RGB input can reach a higher score an all models. To get higher rewards, it is very important to have a good expert. This sounds obvious but the reason behind it is not that straightforward: the neural network is totally unaware of rewards. In other words, the model is just trying to replicate the expert in imitation learning. It’s accuracy increases if it does all the mistakes the expert did.

Training accuracy is correlated with driving performance, but not so strongly. Sometimes I generated agents with lower training accuracies, but higher average rewards. The reason for that is training accuracy is not strongly related to the driving performance. The complex model will lead to overfitting, like the loss nearly decreased to 0 when using the VGG16 model. Also, images with less useless features make convergence faster and allow good results with simpler and faster architectures.

\smallskip

{\sc Acknowledgment:} This work has benefitted enormously from conversations with Hairuo Sun and Eshed Ohn-Bar.

\bibliographystyle{IEEEtran}

\end{document}